\documentclass{article}

\usepackage[preprint, nonatbib]{neurips_2024}

\usepackage[utf8]{inputenc} % allow utf-8 input
\usepackage[T1]{fontenc}    % use 8-bit T1 fonts
\usepackage[table,xcdraw]{xcolor}
\usepackage{hyperref}       % hyperlinks
\usepackage{url}            % simple URL typesetting
\usepackage{booktabs}       % professional-quality tables
\usepackage{amsfonts}       % blackboard math symbols
\usepackage{nicefrac}       % compact symbols for 1/2, etc.
\usepackage{microtype}      % microtypography
\usepackage{xcolor}         % colors
\usepackage{amsmath}        % equations
\usepackage[utf8]{inputenc}
\usepackage{paralist}
\usepackage{todonotes}
\usepackage{CJKutf8}
\usepackage[backend=bibtex,style=numeric,sorting=none]{biblatex} % APA style
\usepackage{authblk}
\usepackage{graphicx}   % 插入图片
\usepackage{subcaption} % 支持子图
\usepackage{float}
\usepackage{wrapfig}
\usepackage{colortbl}

\usepackage{listings}       % add python code
\title{Adaptable and Precise: Enterprise-Scenario LLM Function-Calling Capability Training Pipeline}

\author{Guancheng Zeng\textsuperscript{1}}
\author{Wentao Ding}
\author{Beining Xu}
\author{Chi Zhang}
\author{Wenqiang Han}
\author{Gang Li}
\author{Jingjing Mo}
\author{Pengxu Qiu}
\author{Xinran Tao}
\author{Wang Tao}
\author{Haowen Hu\textsuperscript{2†\thanks{†Corresponding Author}}}
\affil{Digital China AI Research \\ 
\{cenggc\textsuperscript{1}, huhwa\textsuperscript{2}\}@digitalchina.com}

\begin{document}

\maketitle

\begin{abstract}
Enterprises possess a vast array of API assets scattered across various functions, forming the backbone of existing business processes. By leveraging these APIs as functional tools, enterprises can design diverse, scenario-specific agent applications, driven by on-premise function-calling models as the core engine. However, generic models often fail to meet enterprise requirements in terms of computational efficiency, output accuracy, and stability, necessitating scenario-specific adaptation. In this paper, we propose a training pipeline for function-calling capabilities tailored to real-world business scenarios. This pipeline includes the synthesis and augmentation of scenario-specific function-calling data, model fine-tuning, and performance evaluation and analysis. Using this pipeline, we generated 1,260 fully AI-generated samples and 1,035 augmented manually-labeled samples in digital HR agent scenario. The \textbf{Qwen2.5-Coder-7B-Instruct} model was employed as the base model and fine-tuned using the LoRA method on four GPUs with 24GB VRAM. Our fine-tuned model demonstrated outstanding performance in evaluations and practical applications, surpassing \textbf{GPT-4} and \textbf{GPT-4o} in accuracy on the test set. These results validate the reliability of the proposed pipeline for training scenario-specific function-calling models.
\end{abstract}

\section{Introduction}
In the era of generative AI, AI agents are defined as autonomous systems capable of handling tasks independently \cite{yao2022react,xi2023rise}. These agents can not only respond directly to user queries but also perform complex task decomposition and execution by invoking various functional tools \cite{wang2024tdag, jeyakumar2024advancing}. With a vast knowledge base and the ability to interact with external environments, AI agents demonstrate significant potential for deep integration with existing enterprise systems \cite{5331703}. Across industries, enterprises are actively exploring the integration of AI agent systems into real business scenarios \cite{milosevicenterprise}. This is driven by the abundance of API assets within enterprise systems, which, when leveraged as functional tools, enable AI agents to deeply integrate with existing systems, automate workflows across specialized business scenarios, and enhance operational efficiency.

AI agents require LLMs (Large Language Models) as their core reasoning engines to generate instructions for invoking functions and APIs \cite{schwartz2023enhancing}. While both open-source and commercial LLMs perform well in general function-calling tasks given the extensive training data \cite{chung2024scaling,taori2023alpaca}, these models often struggle to provide accurate and stable function-calling instructions in specialized enterprise scenarios. Errors in instruction structure, function selection, and parameter input still occur due to the lack of domain-specific training data. Furthermore, to prevent data leakage \cite{carmody2021ai,}, enterprises typically opt to deploy on-premise LLMs rather than utilizing LLMs served on public cloud platform. However, due to the limited computational resources and lack of experience in training LLMs, it is difficult to implement AI agent effectively for Small and Medium Enterprises. Therefore, there is an urgent need to identify an efficient training pipeline suitable for developing small-scale models in enterprise scenarios, so that SMEs can easily train and deploy their own agent models based on their needs.

In this work, we designed a specialized training pipeline for function-calling capabilities tailored to professional scenarios. This pipeline includes data synthesis and augmentation for scenario-specific function-calling task, SFT (Supervised Fine-Tuning) generic function calling models
using LoRA \cite{hu2021lora}, and comprehensive evaluation and analysis of model performance. Specifically, we used 14 workflow sets and generated 1,260 fully AI-synthesis and 1,035 manually augmented function-calling training samples in a digital HR intelligent agent system, allowing the model to better align with actual business preferences in tool selection and parameter extraction. The \textbf{Qwen2.5-Coder-7B-Instruct} model \cite{bai2023qwen} was used as the base model and fine-tuned using LoRA within approximately five hours on four \textbf{A10} GPUs (24GB VRAM each). The fine-tuned model exhibited exceptional performance, surpassing \textbf{GPT-4} and \textbf{GPT-4o}\footnote{https://chatgpt.com/} in function calling output structural completeness, tool selection accuracy, and parameter input accuracy, both on the test set and in real-world usage. These results demonstrated the effectiveness of our pipeline in enhancing the function-calling capabilities of moderate-size LLMs for professional scenarios under limited computational resources.

In summary, our main contributions are as follows:
\begin{itemize}
    \item We developed a function-calling training pipeline, enabling data synthesis, model fine-tuning, and evaluation for various enterprise-specific tools, significantly improving the performance of generic models in professional business scenario.
    \item We trained a 7B function-calling LLM for the digital HR scenarios. The model exhibited outstanding performance in both evaluations and real-world usage, surpassing GPT-4 and GPT-4o in test set accuracy.
\end{itemize}

The remainder of the paper is organized as follows: Sec.\ref{sec:related-work} refers to related works of data synthesis and model evaluation on Function Calling; 
Sec.\ref{sec:method} describes the pipeline of our work, including initial setting, data synthesis, model training and evaluation; Sec.\ref{sec:exp} contains our experiments as well as results; ablation study and analysis are in Sec.\ref{sec:ablation}.

\section{Related Work} \label{sec:related-work}
\subsection{Function-Calling Data Synthesis}
The introduction of Large Language Models (LLMs) has catalyzed a significant paradigm shift in the field of deep learning \cite{zhao2023survey}. Their ability to produce text that rivals human quality renders them invaluable for generating synthetic data, thereby alleviating the long-standing dilemma of data quality and quantity in Natural Language Processing (NLP) \cite{long2024llmsdrivensyntheticdatageneration,wang2022self,xu2023wizardlm}. 
LLMs have also demonstrated remarkable proficiency in generating function-calling data.\textbf{Toolformer} \cite{schick2024toolformer}, for instance, enhances an LLM's ability to identify retrievable information through function calls, by annotating examples and using augmented data for training, thus enabling the model to effectively determine tool usage. 
Similarly, \textbf{ToolLLaMA} \cite{qin2023toolllm} leverages over 16,000 API instances and datasets generated by ChatGPT to enhance its handling of both simple and complex multi-tool instructions, showcasing robust generalization with novel API documentation. To further simulate real user instructions and enhance the richness and complexity of commands, \textbf{ToolACE} \cite{liu2024toolace} and \textbf{ToolAlpaca} \cite{tang2023toolalpaca} employ multiple LLM agents to simulate multi-turn interactions, generating tool usage examples. Additionally, to ensure data quality, both \textbf{ToolACE} and \textbf{APIGen} \cite{liu2024apigen} implement multiple validation processes to filter data that adhere to correct invocation formats and align the semantic consistency between instruction goals and API outputs.

While LLMs with general Function-calling capabilities are emerging in abundance, there is still no standardized process for fine-tuning Function-calling LLMs in specific professional domains. To address this, we provide a comprehensive workflow for data synthesis, model fine-tuning, and model evaluation tailored to specific business scenarios. This workflow enables LLMs to better understand user instructions and generate appropriate responses by leveraging APIs effectively.

\subsection{Function-Calling Evaluation Metrics}
During model training, the loss function indicates the error for each forward pass, guiding the gradient updates in back-propagation. However, relying solely on loss often fails to capture the model’s task-specific performance. Researchers therefore employ discrete evaluation metrics like code execution success, multiple choice and integer scores to assess performance. The field has developed task-specific evaluation benchmarks such as \textbf{MMLU}, \textbf{CMMLU}, \textbf{C-Eval}, \textbf{GSM8K}, \textbf{ELM}, and \textbf{OpenCompass} \cite{hendrycks2020measuring,li2023cmmlu,huang2024c,cobbe2021training,chan2024mle,2023opencompass}, enhancing model assessment across diverse tasks like code synthesis and commonsense reasoning.

In function-calling studies, numerous evaluation methodologies have been proposed, leading to the emergence of both open-source and proprietary benchmarks. These benchmarks are designed to assess a model's function-calling abilities, often relying on synthetic methods to construct datasets due to the scarcity of function-calling data compared to regular types of data. For instance, \textbf{APIBench} \cite{patil2023gorilla} builds its evaluation data using APIs from platforms such as TorchHub, TensorHub, and Hugging Face to assess models' accuracy and hallucination rates in function calls. For certain task types, there is a greater emphasis on multi-turn and multi-step function-calling. \textbf{ToolBench (ToolEval)} \cite{xu2023tool} collects tool APIs from the RapidAPIHub platform and uses ChatGPT to generate diverse instructions for utilizing these APIs. Similarly, \textbf{AgentBoard} \cite{ma2024agentboard} provides nine categories of tasks and evaluates models' multi-turn agent interaction capabilities using \textbf{ToolOperation} and \textbf{ToolQuery}. The \textbf{BFCL} \cite{berkeley-function-calling-leaderboard} framework introduced a \textbf{Live Dataset} in its V2 version, built from real-time user-contributed function documentation and queries, to evaluate models' ability to operate effectively in dynamic environments. Its V3 version further incorporated multi-step and multi-turn evaluation logic.

For evaluating results, multiple assessment methods have been developed. \textbf{ToolBench} employs GPT-series models to evaluate model performance based on pass rates. On the other hand, benchmarks like \textbf{APIBench} and \textbf{BFCL} utilize \textbf{Abstract Syntax Trees (AST)} to parse the structure of model-generated function calls. This approach enables a multi-faceted evaluation without actually executing the call instructions, allowing for more efficient assessments. Additionally, the \textbf{BFCL} framework offers the \textbf{Exec} method, which evaluates the results of executing the model-generated function-calling instructions, making it more suitable for multi-turn dialogue evaluation. In practical use, researchers typically choose evaluation methods based on their specific needs and available resources.

\section{Methodology} \label{sec:method}
\subsection{Overview}
\begin{figure}[t]
    \centering
    \includegraphics[width=1\linewidth]{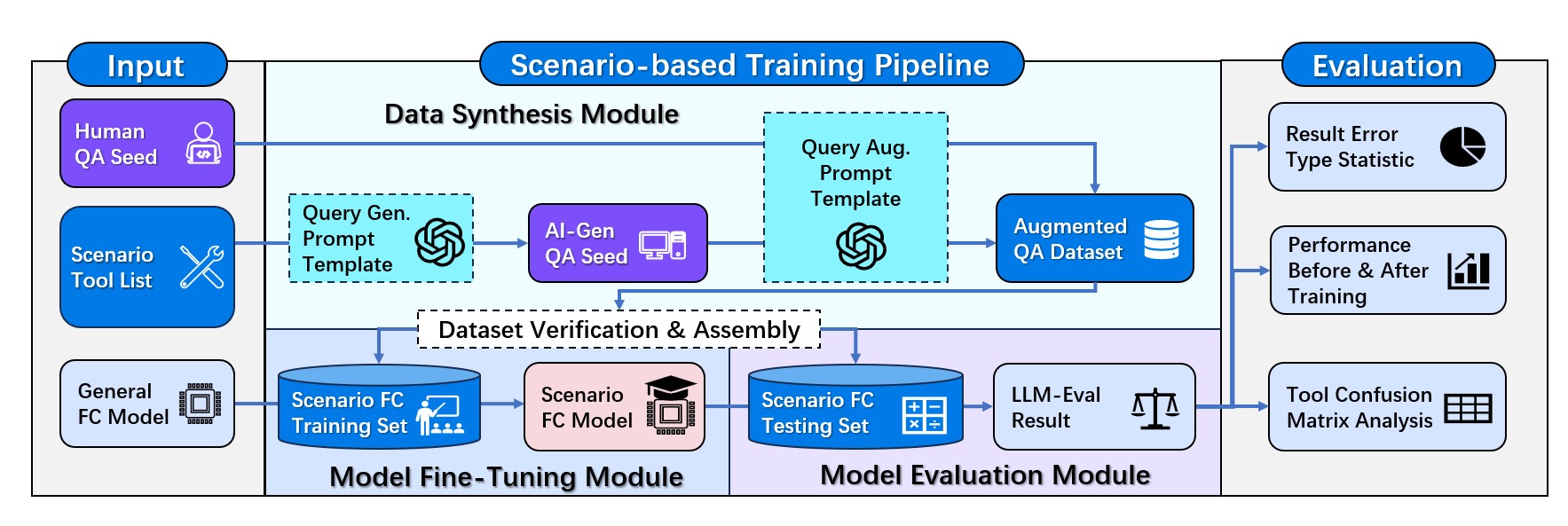}
    \caption{The Overall Training Pipeline of Enterprise-Scenario Function-Calling LLM}
    \label{fig:pipeline}
\end{figure}

In enterprise environments, characterized by relatively high labor costs, it is crucial to minimize manual involvement \cite{hutter2019automated}. Therefore, we designed a highly automated training pipeline for function-calling capabilities in enterprise models tailored to such scenarios, as shown in Figure \ref{fig:pipeline}.

This pipeline consists of three main modules:
\begin{itemize}
    \item \textbf{Data Synthesis Module:} This module generates user questions based on function tool information and creates corresponding function-calling instructions. It enhances and filters the generated questions to further improve the quantity and quality of the data \cite{wei2022chain}. The resulting data is then assembled and divided into fine-tuning training sets and evaluation sets \cite{brown2020language}.
    \item \textbf{Model Fine-Tuning Module:} Using the training set, this module fine-tunes a generic function-calling model via \textbf{LoRA} \cite{hu2021lora} and integrates the resulting weights back into the original model \cite{pfeiffer2020adapterfusion}.
    \item \textbf{Model Evaluation Module:} After training, the model undergoes AST-based evaluation to assess its performance \cite{allamanis2017learning}. Additionally, the module performs a confusion matrix analysis of the model's function selection distribution, identifying its precision and areas of confusion for different function tools \cite{sokolova2009systematic}.
\end{itemize}
\subsection{Initial Settings}
\subsubsection{Generic Model}
In enterprise scenarios, computational resources are often limited, and small to medium-sized enterprises typically struggle to support the fine-tuning of large-parameter models \cite{houlsby2019parameter,han2015deep}. As a result, smaller-parameter models are required. Furthermore, these models need to possess general function-calling capabilities to enable the transfer of their abilities from general scenarios to enterprise-specific contexts \cite{ruder2019transfer}. Previous research has demonstrated that models with 7B parameters can effectively handle general function-calling tasks, achieving high accuracy levels in real-world applications 
 \cite{kaplan2020scaling}.
    
\subsubsection{Scenario API Tools} 
Depending on the task, models can utilize various types of tools, including APIs, algorithms, code workflows, operational pipelines, and even other models. The model must be able to select these tools correctly based on their format and provide the appropriate input parameters \cite{gao2020making}. To enable stable data synthesis and tool usage, users need to supply the following information about each function tool: \textbf{Tool Name}, \textbf{Tool Description}, \textbf{Parameter Names}, \textbf{Parameter Descriptions}, \textbf{Parameter Data Types}, \textbf{Parameter Necessity} \cite{wang2022self}. Furthermore, for parameters that require high accuracy, users can enhance the stability of tool invocation by providing specific examples and default values for the input parameters \cite{brown2020language}. Descriptions of function tools and their parameters must be provided by the user, which should clearly convey the tool's functionality and parameter requirements.  The higher the quality of these descriptions, the better the quality of the data generated, which in turn enhances the outcomes of model training \cite{li2023quantity}.

\subsubsection{Human-Annotated Seed Data} 
To ensure that the data closely align with real-world scenarios, the pipeline could initially incorporate a small amount of manually annotated data as seed data for the data synthesis module. Compared to fully AI-generated data, seed data annotated by business experts is better aligned with actual question-and-answer patterns, improving the quality of the generated data. This, in turn, enhances the model's stability and accuracy \cite{snow2008cheap}.

During manual annotation, several key aspects must be considered:
\begin{compactitem}
    \item \textbf{Diversity}: Questions should exhibit diversity in expression, including variations in sentence structure, content, and phrasing \cite{wei2019eda}.
    \item \textbf{Uniqueness}: Each data entry should be unique in terms of questions and parameters, avoiding repetition of names (e.g., people, places, departments, or projects) \cite{batini2009methodologies}.
    \item \textbf{Scale}: The seed data should be of sufficient size, with its volume proportionally allocated according to the importance of the function \cite{kaplan2020scaling}.
    \item \textbf{Consistency}: The internal logic of the seed data must remain consistent, with uniform output formats and standardized parameter names for the same functionality \cite{wang1996beyond}.
\end{compactitem}
By adhering to these principles, the seed data ensures a high-quality foundation for data synthesis and model training, optimizing overall performance in specific application scenarios.

\subsection{Data Synthesis}
In this process, questions are first generated based on the descriptions of scenario-specific function tools\cite{liu2024apigen,wei2019eda}. These questions, along with a small number of manually annotated question-answer pairs, are used as seed data for data augmentation \cite{batini2009methodologies}. The augmented data are then utilized as the training set to fine-tune the model \cite{gururangan2020don}.

\subsubsection{Generating Seed Questions Based on API Description} 
During model fine-tuning, it has been observed that the diversity, quality, and quantity of the dataset directly influence training outcomes \cite{liu2024apigen}. Diversity is critical for the model's performance on unseen tasks or instructions \cite{brown2020language,gao2020pile}, quality impacts its accuracy on known tasks \cite{gururangan2020don}, and quantity effectively prevents the model from over-fitting to specific tasks. When synthesizing function-calling training datasets, the design of seed questions plays a pivotal role in determining the model's final performance. These seed data must cover as many task scenarios as possible while adhering to the logical boundaries of the tool's functionality. The generated questions must not deviate from the purpose described in the tool \cite{brown2020language,gururangan2020don}.

To achieve this, we designed various customization prompt templates in the pipeline, incorporating elements such as role settings, generation rules, and output formats. These templates must meet the following requirements:
    \begin{itemize}
        \item \textbf{Role Setting}: Define the identity of the user asking the question, such as a salesperson, domain expert, or engineer, ensuring that the roles are diverse, clearly defined, and logically aligned with the scenario \cite{ouyang2022training}.
        \item \textbf{Generation Rules}: Specify requirements for data synthesis, such as the number of questions generated per prompt, question length, required content, content to avoid, and language of the generated questions \cite{gao2020pile,raffel2020exploring}.
        \item \textbf{Output Format}: Standardize the output format to facilitate automated data counting and dataset assembly. Unnecessary prefixes and suffixes in the generated output should be minimized \cite{gururangan2020don,ouyang2022training}.
    \end{itemize}
While research indicates that small-parameter models can produce more data with the same resource consumption, potentially enhancing training outcomes \cite{bansal2024smaller,bender2020climbing}, their limited instruction-following capabilities make them unsuitable for complex prompt templates required in specialized scenarios. As a result, such models often produce data with formatting issues and high redundancy. Therefore, for generating initial seed data in professional contexts, high-performance models remain essential \cite{raffel2020exploring,bender2020climbing}.

\subsubsection{Data Augmentation Based on Seed Questions} 
The pipeline, starting from seed questions, broadens scenario coverage through data augmentation. This method, compared to direct annotation, lowers resource consumption. It also proves more stable than generating all data simultaneously and reduces the chance of producing duplicates. Data augmentation, like data synthesis, depends on LMs and prompt templates. However, the diversity required in augmentation directions necessitates a broader array of templates. Overall, our pipeline employs four data augmentation strategies: replacement, rewriting, simplification, and error introduction. The definition and function of these four augmentation strategies are shown in Table \ref{tab:Augmentation}. Samples of augmentation prompt templates in digital HR scenario can be found in \ref{sec:apx_b}.

\begin{table}[htbp]
\centering
\caption{Definition and Function of Augmentation Prompt Strategies}
\label{tab:Augmentation}
\begin{tabular}{ll}
\toprule
Augmentation Strategies &
  Definition and Function \\ \hline \vspace{-5pt}\\
Replacement &
  \begin{tabular}[c]{@{}l@{}} Non-essential elements in the seed questions are replaced without \\ affecting the selection of function-calling instructions or parameter \\ inputs.
  \\\end{tabular} \\\vspace{-5pt} \\\hline \vspace{-5pt}\\
Rewriting &
  \begin{tabular}[c]{@{}l@{}} The language structure of the seed questions is rearranged, altering \\ the grammar and phrasing of the queries while retaining their original \\ meaning.
  \\\end{tabular} \\\vspace{-5pt} \\\hline \vspace{-5pt}\\
Simplification &
  \begin{tabular}[c]{@{}l@{}} Non-essential content and conjunctions in the seed questions are \\ removed, shortening the original sentence into a concise query while \\ maintaining key information. This simulates real-world user queries \\ that often consist of only essential keywords.
  \\\end{tabular} \\\vspace{-5pt} \\\hline \vspace{-5pt}\\
Error Introduction &
 \begin{tabular}[c]{@{}l@{}} Errors such as misspellings, repetitions, or deletions of non-critical \\ content are added to the seed questions to help the model adapt to \\ occasional errors in user input.
 \\\end{tabular}\\\vspace{-8pt}\\ \bottomrule
\end{tabular}
\end{table}

In practice, replacement and rewriting are the most commonly used augmentation methods, while simplification and error introduction are employed less frequently. Multiple augmentation methods can be applied to the same seed question. Note that error introduction should be based on already-augmented questions to avoid excessive similarity to the original seed questions.

\subsubsection{Generating Function Calling Instructions Based on Questions} 
After generating the questions, function-calling instructions can be created based on these questions and included in the dataset as labels and answers. The generation of instructions primarily relies on the extraction of key parameters \cite{schick2024toolformer}. At this stage, it is generally necessary to use high-performance models for parameter extraction to ensure consistency between the extracted function-calling parameters and the questions \cite{yao2022react}. Once the parameters are obtained, they need to be assembled into function-calling instructions in accordance with the format required by the training and testing datasets \cite{berkeley-function-calling-leaderboard,zheng2024llamafactory}. This process involves integrating the parameters with the tools relevant to the problem's context. Consequently, all augmented questions generated from the same seed question use the same function-calling instructions.

\subsubsection{Data Validation and Assembly} 
Since certain parameters in specific problems are generated synthetically by the model and may not accurately reflect real-world conditions, the final augmented training data must undergo a validation process \cite{howard2018universal}. This involves checking their structural integrity and verifying the accuracy of parameter assignments.
    
After data validation, the remaining dataset is divided into training and validation sets while ensuring that all usage scenarios of the tools are covered \cite{kohavi1995study}. Additionally, the data format must be adjusted to align with the requirements of the training and evaluation frameworks. Typically, each data entry should include the user's question, the model's tool-calling instructions, and a list of available tools, formatted in JSON. Commonly used formats include \textbf{shareGPT}, \textbf{Alpaca}, and \textbf{OpenAI} formats.
    
Regarding the list of available tools, it must align with the task coverage of the scenarios and provide all the required tools \cite{sun2019fine}. The number of available tools and the length of their descriptions should be kept within a certain range to avoid exceeding the model's context window size and to reduce the complexity of tool selection \cite{vaswani2017attention}. During data assembly, the order of the tool list in the training data should be randomized to prevent the model from over-fitting to specific sequences of tool combinations \cite{srivastava2014dropout}.

\subsection{Model Fine-tuning}
\subsubsection{LoRA}
To prevent over-parametrization during scenario-specific task adaptation and to save computational resources, the process employs the \textbf {LoRA} (Low-Rank Adaptation) fine-tuning approach \cite{hu2021lora}. This reduces the number of parameters required for adaptation while effectively preserving the base model's general capabilities. During training, the model’s context length must always exceed the length of the training data, avoiding truncation of longer data entries to prevent structural disruption, which could affect training performance \cite{vaswani2017attention}. The number of training epochs should not be excessively short or long, as this may result in under-fitting or over-fitting on the training dataset. 
    
We use \texttt{LLaMA-Factory} \cite{zheng2024llamafactory} as the training framework in this pipeline, which supports the adaptation of various models and provides diverse fine-tuning methods. The training is primarily implemented using \texttt{PEFT} library\footnote{https://huggingface.co/docs/peft/en/index}. Additionally, the \texttt{DeepSpeed-ZeRO3} \cite{rajbhandari2020zeromemoryoptimizationstraining} optimization method is applied to accelerate training. This approach distributes the optimizer, gradients, and parameters across GPUs, significantly reducing memory consumption and enabling more efficient utilization of GPU resources.

\subsubsection{Merging LoRA Adapters} 

After completing LoRA fine-tuning, the LoRA adapters are merged to the base model to facilitate further fine-tuning tasks. Additionally, the pipeline includes a  \textbf {multi-adapter merging strategy}, which encompasses various processing methods such as \textbf{linear combination}, \textbf{SVD} \cite{stoica2024model}, and \textbf{parameter concatenation} \cite{peft}. These strategies enable the merging of weight matrices from different LoRA adapters prior to integration, addressing training scenarios such as multi-task fine-tuning, and data feedback fine-tuning, supports continuous updates to the model after deployment. This is crucial for business scenario as data often follows a life-cycle with continuous replacement of updated data. Furthermore, with this multi-LoRA adapter merging approach, the influence of different training data can be encapsulated within separate LoRA adapters. By selecting specific LoRA adapters, it becomes easier to mitigate the impact of certain data cycles or tasks on the model’s performance, offering enhanced flexibility in managing the model's behavior and adaptability.

\subsection{Model Evaluation}
\subsubsection{AST Evaluation} 
To effectively verify the correctness of the function-calling instructions generated by the model, evaluation must address multiple aspects, including \textbf{structural completeness of the instructions}, \textbf{accuracy of function selection}, and \textbf{accuracy of parameter inputs}. The \textbf{AST} (Abstract Syntax Tree) evaluation method enables a step-by-step parsing of the model's output structure, allowing the correctness of function-calling instructions to be assessed without actually executing the function tools \cite{aho2007compilers}. This capability makes AST-based evaluation particularly useful for assessing AI-generated function-calling instructions that contain fictitious parameters, a scenario where \textbf{Exec} (Execution) methods often struggle. The Exec method requires actual execution of the model’s output instructions and evaluates correctness based on the returned results \cite{chen2021evaluating}. However, instructions with fictitious parameters—such as nonexistent IDs, names of people, places, or companies—cannot be executed, especially for query-based API tools. In addition, since AST evaluation does not involve actual API execution, it is not constrained by API response times, enabling faster verification of model outputs \cite{berkeley-function-calling-leaderboard}.

In our pipeline, we adopted the open-source \textbf{BFCL} evaluation framework as the primary benchmark for our practical experiments, modifying it to enhance its suitability for assessing scenario-specific function-calling ability evaluation \cite{berkeley-function-calling-leaderboard}. This framework provides a rapid and comprehensive evaluation system, including AST-based instruction parsing methods and error analysis features. These features facilitated advanced analyses of the model's function-calling outputs, enabling deeper insights into the causes of errors and refining the evaluation process.

\subsubsection{Confusion Matrix for Tool Selection} 
In addition to exploring the causes for errors in function-calling instructions, tool selection can be evaluated as a multi-class classification task, given that the model selects a specific tool from a list of available tools. This evaluation employs a confusion matrix to analyze the accuracy of tool selection \cite{powers2020evaluation}. Specifically, each function tool category is treated as an independent class, and a confusion matrix is constructed where rows represent the actual tool categories and columns represent the model’s predicted categories. This approach facilitates the computation of the number of instances where the model accurately predicted a tool for each real tool’s usage scenario \cite{sokolova2009systematic}.

Using the confusion matrix, the usage of a target tool is categorized as a positive instance, whereas the usage of alternative tools is categorized as a negative instance. This enables the calculation of \textbf{True Positives (TP)}, \textbf{True Negatives (TN)}, \textbf{False Positives (FP)}, and \textbf{False Negatives (FN)} for each tool in its respective usage scenario. These metrics, along with their definitions, are listed in Table \ref{tab:Symbol}.

\begin{table}[htbp]
\centering
\caption{Value and Meanings of Tool Selection Confusion Matrix}
\label{tab:Symbol}
\begin{tabular}{cl}
\toprule
Value & Meaning in Tool Selection \\
\hline \vspace{-7pt} \\
TP &  Selected the target tool in the target tool scenario \\
TN &  Selected another tool in a non-target tool scenario \\
FP & Incorrectly selected the target tool in a non-target tool scenario \\
FN & Incorrectly selected another tool in the target tool scenario \\
\bottomrule
\end{tabular}
\end{table}

Based on these values, the following key performance metrics can be derived to evaluate the model’s tool selection capability:
    \begin{itemize}
        \item \textbf{Precision}: The proportion of correct predictions among all instances where the model predicts the use of a specific tool.
        \begin{equation}
            \text{Precision} = \frac{\text{TP}}{\text{TP} + \text{FP}}
        \end{equation}
        \item \textbf{Recall}: The proportion of correct tool selections in the total number of instances where the tool is actually used in the scenario.
        \begin{equation}
            \text{Recall} = \frac{\text{TP}}{\text{TP} + \text{FN}}
        \end{equation}
        \item \textbf{F1 Score}: The harmonic mean of precision and recall, providing a comprehensive assessment of the model’s performance in selecting a specific tool.
        \begin{equation}
            F_1 = 2 \cdot \frac{\text{Precision} \cdot \text{Recall}}{\text{Precision} + \text{Recall}}
        \end{equation}
    \end{itemize}
    
These metrics reflect the model's capability to select tools and can be employed to compare its performance before and after fine-tuning, as well as against high-performance models.

Furthermore, conducting confusion matrix analysis with evaluation results from high-performance generic models in zero-shot setting can provide insights into the correspondence between tool descriptions and user queries, as well as the confusion degree between tools. Tools with low precision rate are more likely to interfere with the selection of other tools, while tools with low recall rate are more likely to be affected by other tools. Based on these observations, practitioners can optimize the functional scope and descriptive information of tools prior to data synthesis. This ensures that each tool has clearly differentiated functionality and that its description aligns with its logical use case within the given scenario, which helps reduce tool confusion and improves the overall accuracy of tool selection by the model.

\section{Experiments} \label{sec:exp}
\subsection{Experimental Background}
In our experiment, we tested our pipeline in a digital HR intelligent agent scenario within a large corporate group, involving over 8,000 employees. Users could interact with the intelligent agent in Chinese, inquiring about information related to the company's employees and departments. The experiment provided a total of 14 specialized workflows with distinct functionalities, each encapsulated in the format of function tools. The model needs to interpret user's queries accurately, pass them as parameters to the workflows. Workflow will automatically query the relevant data from the database, summarize the results, and then deliver them back to the users.

\subsection{Experimental Setup}
\subsubsection{Data Synthesis} 
In this experiment, the training data were synthesized in two stages: seed question generation and data augmentation. These tasks were completed using GPT-4 equipped with appropriate prompt templates, aiming to maximize the quality of the generated data. Details of the prompt templates can be found in \ref{sec:apx_a} and \ref{sec:apx_b}.

During the data synthesis process, for each of 14 function tools (scenario workflows), 10 seed questions were generated, leads to 140 seed questions in total. These seed questions underwent data augmentation, each produce 10 enhanced question, which gave us 1400 augmented data instances. From these augmented data, we finally selected 1260 data instances as training data, 90 instances for each tools.

\begin{figure}
    \centering
    \includegraphics[width=1\linewidth]{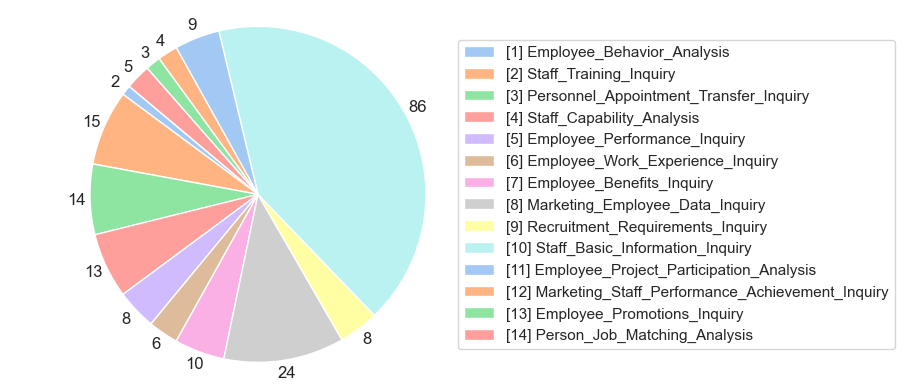}
    \caption{Data Volume and Distribution of 14 Digital HR Scenario Tools}
    \label{fig:data-distribution}
\end{figure}

Furthermore, we incorporated 207 manually annotated data instances to compare the performance improvements between two types of seed data. From these 207 manually annotated seeds, we generated a total of 1,035 augmented training data instances, with 5 augmentations for each seed instance. Notably, to align with real-world scenarios usage, the data distribution of the 207 manually annotated data instances across the 14 workflows was uneven. The distribution details are shown in Figure \ref{fig:data-distribution}. 

Among the workflows, \textbf{[10]Staff Basic Information Inquiry} accounts for more than one-third of the total data volume, as it is primarily used for querying basic user information. Additionally, some workflows share similar functions and descriptions, such as \textbf{[12]Marketing Staff Achievement Inquiry} and \textbf{[8]Marketing Employee Data Inquiry}, as well as \textbf{[4]Staff Capability Analysis}, \textbf{[14]Person Job Matching Analysis}, and \textbf{[10]Staff Basic Information Inquiry}. These similarities increase the difficulty for the model to select the correct tool during function calls, leading to potential errors in tool selection. As shown in Figure \ref{fig:confusion_matrix}, both Qwen (b) and GPT-4 (c) encounter tool selection errors in certain workflows under zero-shot conditions, highlighting the complexity of this task and the necessity of fine-tuning.

Regarding data formatting, the training data was organized in the ShareGPT format. During assembly, order of the tool list was randomized to prevent the model from over-fitting to specific tool sequence patterns. For evaluation, the data followed the BFCL\_V3 framework's built-in format.

\subsubsection{Model Fine-tuning} 
In this experiment, we selected Qwen2.5-Coder-7B-Instruct as the base model and performed LoRA fine-tuning with four NVIDIA A10 GPUs (24GB VRAM each). Under the default hyper-parameter settings, the batch size was set to 1, with a gradient accumulation step of 16 to simulate large-batch gradient updates under limited computational resources. The learning rate warm-up ratio was set to 0.1, with a peak learning rate of $8.0 \times 10^{-5}$, followed by a cosine learning rate decay to gradually reduce the learning rate after reaching its peak. 

The training process used bf16 precision and ran for 10 iterations. For testing, checkpoints were typically selected between the $7^{th}$ and $10^{th}$ iterations. During ablation experiments, the same number of data iterations was maintained to ensure consistency. In the LoRA configuration, the rank size $r$ was set to 8, the scaling factor $\alpha$ was set to 16, and the dropout rate was set to 0. LoRA fine-tuning was applied to all modules in the model.

During fine-tuning, we used different datasets to train the model, including:
\begin{itemize}
    \item \textit{DHR\_train\_1}: Containing 1,260 AI-augmented seed data instances.
    \item \textit{DHR\_train\_2}: Containing 1,035 human-augmented seed data instances.
    \item \textit{DHR\_train\_3}: A combined dataset containing 1,260 AI-augmented and 1,035 human-augmented seed data instances, totaling 2,295 data instances.
\end{itemize}
The average final token length of each data entry exceeded 3,000 tokens across these datasets.

The experiment compared the impact of different datasets on the model's performance after fine-tuning. Furthermore, we examined the performance disparities of models trained on a dataset that combined AI-generated and human-augmented data under different training conditions, and compared their effectiveness with that of generic models.

\subsubsection{Model Evaluation} 
In this experiment, we evaluated the function-calling instructions generated by the model using the AST method. A total of 207 manually annotated seed questions were selected as the \textit{DHR\_test\_A} test set to compare the effects of different hyper-parameter settings and mixed data on model fine-tuning. Additionally, we designed 135 distinct questions as the \textit{DHR\_test\_B} test set to prevent evaluation bias caused by potential over-fitting to seed data.
    
After obtaining the model's output and its overall accuracy score, we further analyzed the distribution of error types in certain test scenarios. These error types included \textbf{Structure Errors (SE)}, \textbf{Tool Errors (TE)}, \textbf{Parameter Errors (PE)}. These errors are calculated in certain orders during evaluation: Function-calling instructions with structural parsing errors could not be evaluated for tool selection or parameter input accuracy, while instructions with tool selection errors could not be evaluated for parameter input accuracy. The calculation formulas are expressed in the equations below:

\begin{equation}
P(\text{Structural Completeness Rate}) = \frac{N(\text{Test Set})-N(\text{SE})}{N(\text{Test Set})}
\end{equation}

\begin{equation}
P(\text{Tool Selection Acc.}) = \frac{N(\text{Test Set})- N(\text{SE}) -N(\text{TE})}{N(\text{Test Set}) - N(\text{SE})}
\end{equation}

\begin{equation}
P(\text{Parameter Filling Acc.}) = \frac{N(\text{Test Set})- N(\text{SE}) - N(\text{TE}) -N(\text{PE})}{N(\text{Test Set}) - N(\text{SE}) - N(\text{TE})}
\end{equation}

In addition, we conducted confusion matrix evaluations for the fine-tuned model trained on the \textit{DHR\_train\_3} dataset. We calculated Precision, Recall, and F1 scores for each workflow invoked by the models, providing detailed insights into the model’s performance in workflow selection.

\subsection{Experimental Results}

\begin{table}[]
\centering
\caption{Training Results of Experiments in Digital HR Scenario. The models DHR\_train\_1\_ft, DHR\_train\_2\_ft, and DHR\_train\_3\_ft were trained using their respective datasets.}
\label{tab:training-results}
\begin{tabular}{llllll}
\toprule
 &
  \textit{DHR\_test\_A} &
  \textit{DHR\_test\_B} &
  \begin{tabular}[c]{@{}l@{}}Structural \\ Completeness\\ Rate (\%)\end{tabular} &
  \begin{tabular}[c]{@{}l@{}}Tool \\ Selection \\ Accuracy (\%)\end{tabular} &
  \begin{tabular}[c]{@{}l@{}}Parameter \\ Filling\\ Accuracy (\%)\end{tabular} \\ \hline
\rule{-3pt}{15pt}
\begin{tabular}[c]{@{}l@{}}Qwen2.5-Coder-\\ 7B-Instruct\end{tabular} & 22.7          & 28.1          & 92.8         & {\color[HTML]{333333} 82.8} & 29.6         \\
GPT-4                                                                & 79.2          & 88.1          & 99.0         & 95.1                        & 89.2         \\
GPT-4o                                                               & 32.9          & 34.1          & 94.2         & 52.3                        & 66.7         \\
DHR\_train\_1\_ft                                                    & 85.5          & 89.6          & 100          & 85.5                        & 100          \\
DHR\_train\_2\_ft                                                    & 94.7          & 94.1          & 100          & 94.7                        & 100          \\
DHR\_train\_3\_ft                                                    & \textbf{97.6} & \textbf{95.6} & \textbf{100} & \textbf{97.6}               & \textbf{100} \\ 
\bottomrule
\end{tabular}
\end{table}

Tab.\ref{tab:training-results} presents the training results of our experiments. The models DHR\_train\_1\_ft, DHR\_train\_2\_ft, and DHR\_train\_3\_ft were trained using their respective datasets. As shown, DHR\_train\_1\_ft and DHR\_train\_3\_ft significantly outperformed the baseline models Qwen2.5-Coder-7B-Instruct and GPT-4o across both test sets.
To further analyze the experimental results, we categorized the error types for different models on the \textit{DHR\_test\_A} test set, with the findings summarized in Tab.\ref{tab:training-results}. The analysis revealed substantial improvements in all aspects after fine-tuning, including the completeness of function-calling structures, the accuracy of tool selection, and the accuracy of parameter filling. After fine-tuning, the models achieved complete accuracy in output structure completeness and parameter filling accuracy.
In terms of tool invocation accuracy, DHR\_train\_1\_ft surpassed both the base model and GPT-4o, DHR\_train\_2\_ft reached the performance level of GPT-4, and DHR\_train\_3\_ft exceeded GPT-4. These findings underscore the effectiveness of fine-tuning for specific scenarios, demonstrating that even small-parameter models can be effectively optimized for specialized tasks, thereby narrowing the performance gap with larger, resource-intensive LLMs.

\subsection{Confusion Matrix Evaluation Results}
\begin{figure}[h]
    \centering
    % 第一个子图
    \begin{subfigure}[b]{0.32\textwidth}
        \includegraphics[width=\textwidth]{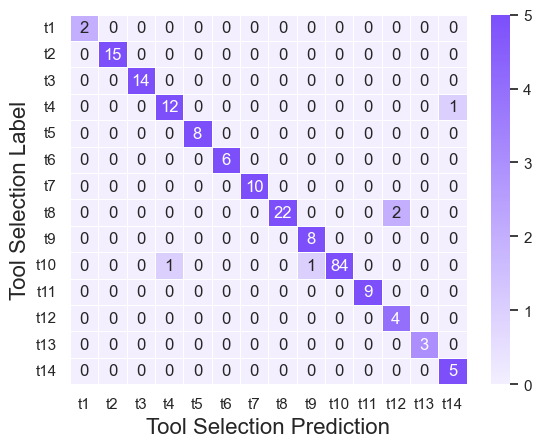}
        \caption{DHR-train-3-ft}
        \label{fig:subfig1}
    \end{subfigure}
    \hfill
    % 第二个子图
    \begin{subfigure}[b]{0.32\textwidth}
    \includegraphics[width=\textwidth]{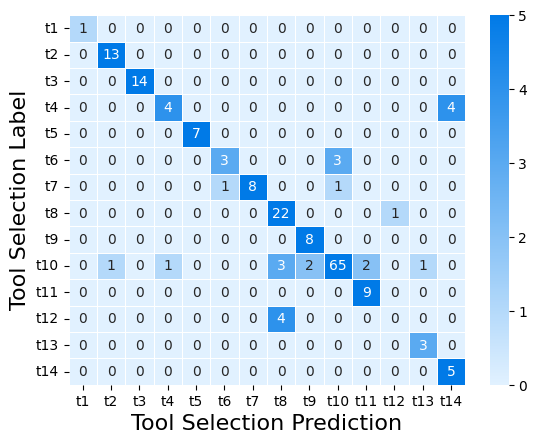}
        \caption{Qwen2.5-Coder-7B-Instruct}
        \label{fig:subfig1}
    \end{subfigure}
    \hfill
    % 第三个子图
    \begin{subfigure}[b]{0.32\textwidth}
    \includegraphics[width=\textwidth]{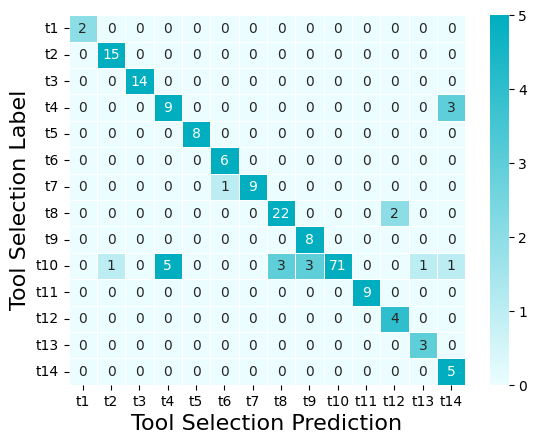}
        \caption{GPT-4}
        \label{fig:subfig1}
    \end{subfigure}
    \hfill
    \caption{Tool Selection Confusion Matrix Heatmap}
    \label{fig:confusion_matrix}
\end{figure}

\begin{figure}[h]
    \centering
    \includegraphics[width=0.9\linewidth]{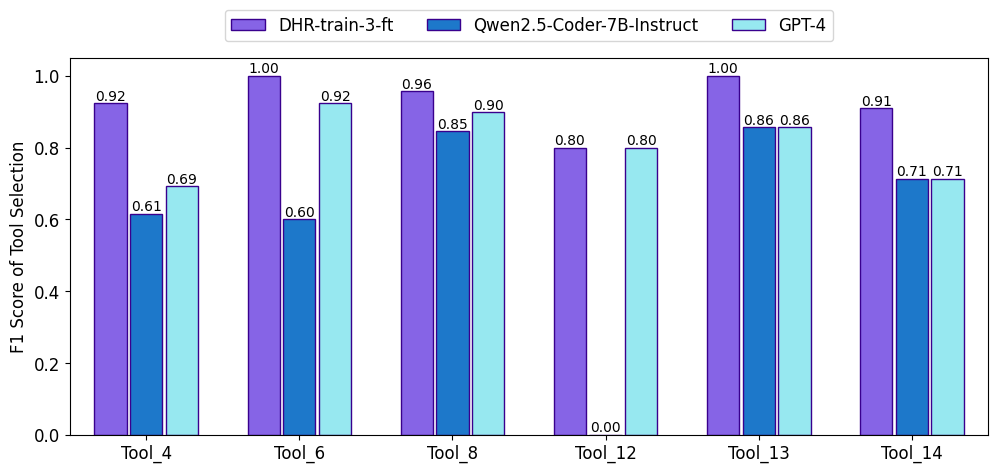}
    \caption{F1 Score Comparison of Different Models across Various Tools. For \textbf{[12] Marketing Staff Achievement Inquiry}, the original model failed all four test instances, thus we use zero to represent its F1 score.}
    \label{fig:F1-score}
\end{figure}

To verify the rationality of these workflow designs and test sets, we also analyzed the performance of the Qwen2.5-Coder-7B-Instruct model after fine-tuning on the \textit{DHR\_test\_A} test set. The results were represented using confusion matrices in Fig.\ref{fig:confusion_matrix} and relevant performance metrics in Fig.\ref{fig:F1-score}. 

The results demonstrate significant improvements in tool selection accuracy after fine-tuning. Compared to the origin model and GPT-4, our fine-tuned model not only achieved near-perfect accuracy in selecting distinctly different tools but also demonstrated improved differentiation between similar tools. Notably, the model showed enhanced F1 score for the three error-prone tools: \textbf{[4] Employee Capability Analysis}, \textbf{[6] Employee Work Experience}, and \textbf{[14] Job Matching}, indicating a comprehensive enhancement of the model's function selection capabilities within the scenario. 

\section{Ablation Study} \label{sec:ablation}
To further investigate the factors affecting the model’s training performance, we conducted a series of in-depth ablation experiments. These experiments involved adjustments to the base model, the composition of training data, data length, and the selection of training hyper-parameters. The tests were conducted on the \textit{DHR\_test\_A} test set, with the results and analyses detailed below:

\subsection{Comparison of Base Models after Training}
First, we explored the impact of different base models on training performance. The experiments focused primarily on the Qwen series models, which were trained using the \textit{DHR\_train\_3} dataset. 

Tab.\ref{tab:compare-base-models} shows that \textbf{Qwen2.5-Coder-7B-Instruct} performed the best and is the most suitable base model for this scenario. The Coder model excels in generating function-calling data formats, as its training set includes a higher proportion of code and structured data, which enhances its performance in the function-calling tasks. Qwen1.5-7b-Chat also achieved very good results, likely due to the DPO (Direct Preference Optimization) training it underwent in later stages, making it more adaptable for fine-tuning in downstream tasks \cite{rafailov2024direct}. In contrast, Qwen2.5-7B, as a pre-trained model without fine-tuning on tasks, did not perform as well as the instruction-tuned models.

\begin{table}[h]
\centering
\caption{Impact of Different Base Models on Training Performance. Qwen2.5-Coder-7B-Instruct performed the best in Qwen series and is the most suitable base model for this scenario.}
\label{tab:compare-base-models}
\begin{tabular}{lllll}
\toprule
 &
  \begin{tabular}[c]{@{}l@{}}Structural\\ Completeness\\ Rate (\%)\end{tabular} &
  \begin{tabular}[c]{@{}l@{}}Tool\\ Selection\\ Accuracy (\%)\end{tabular} &
  \begin{tabular}[c]{@{}l@{}}Parameter\\ Filling\\ Accuracy (\%)\end{tabular} &
  \textit{DHR\_test\_A} \\ \hline
\rule{-3pt}{15pt}
\begin{tabular}[c]{@{}l@{}}Qwen2.5-Coder-\\ 7B-Instruct\end{tabular} &
  \textbf{100} &
  \textbf{97.60} &
  \textbf{100} &
  \textbf{97.60} \\
Qwen2.5-7B-Instruct & 100   & 96.60 & 100   & 96.60 \\
Qwen2-7B-Instruct   & 100   & 90.80 & 100   & 90.80 \\
Qwen1.5-7b-Chat     & 100   & 97.10 & 100   & 97.10 \\
Qwen2.5-7B          & 94.20 & 85.60 & 44.30 & 35.70 \\ \bottomrule
\end{tabular}
\end{table}

\subsection{Comparison Between AI-Generated and Manually Annotated Seed Data}

\begin{wraptable}{r}{0.6\linewidth}
\centering
\caption{Impact of Different Dataset on Training Performance. Each dataset was randomly sampled to include only 1,000 data instances for fine-tuning.}
\label{tab:dataset-type}
\begin{tabular}{lll}
\toprule
Datasets      & \textit{DHR\_test\_A} & \textit{DHR\_test\_B} \\ \hline
\rule{-3pt}{12pt}
\textit{DHR\_train\_1\_1000} & 83.09        & 87.41        \\
\textit{DHR\_train\_2\_1000} & \textbf{95.17}        & 94.81        \\
\textit{DHR\_train\_3\_1000} & 92.27        & \textbf{96.30}        \\ \bottomrule
\end{tabular}
\end{wraptable}

We used Qwen2.5-Coder-7B-Instruct as the base model for subsequent ablation experiments. To compare the impact of different training datasets on model performance, we conducted fine-tuning tasks using the \textit{DHR\_train\_1}, \textit{DHR\_train\_2}, and \textit{DHR\_train\_3} datasets. Each dataset was randomly sampled to include only 1,000 data instances for fine-tuning. The results, shown in Table \ref{tab:dataset-type}, indicate that the DHR\_train\_3\_1000\_ft model achieved particularly significant improvements in both the \textit{DHR\_test\_A} and \textit{DHR\_test\_B} benchmark tests. This demonstrates that training the model using a combination of AI-generated and manually annotated seed data yields better results. One possible explanation is that the AI-generated data expanded the coverage of workflows that were underrepresented in the manually annotated seed data, thereby improving the model's performance in those areas, and preventing the model from over-fitting to several high-frequency tools. The proportion of corresponding data has been increased, which has played a role in balancing the data set.

\subsection{Comparison of Training Data Cutoff Length}
\begin{wrapfigure}{r}{0.6\linewidth} 
    \centering
    \includegraphics[width=0.9\linewidth]{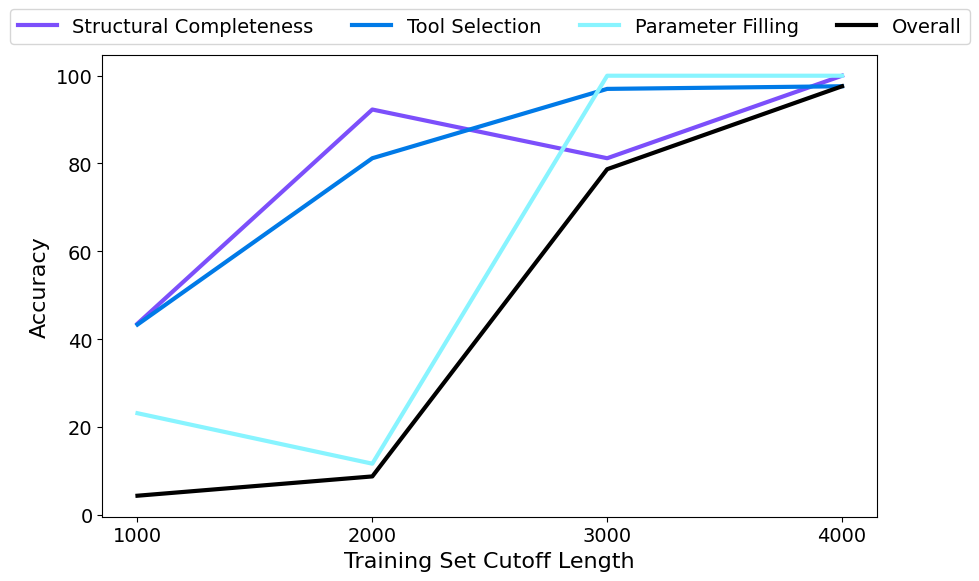}
    \caption{Comparison of Model Performance with Different Training Data Cutoff Length}
    \label{fig:cutoff-length}
\end{wrapfigure}

In fine-tuning tasks involving unstructured natural language, the fine-tuning framework offers a data cutoff method. By setting a maximum input length for training data, this method processes the data by truncating the parts that exceed the defined length, retaining only the portion within the maximum input limit \cite{zheng2024llamafactory}. This approach helps prevent excessive consumption of GPU memory. For tasks that use natural language as training data, truncating text is a common practice, as truncating some content typically does not significantly affect model performance. However, since function-calling tasks rely heavily on structured data, truncating training data could have a more substantial impact. To investigate the effect of cutoff length on structured data, we trained models with various cutoff lengths. The results are shown in Figure \ref{fig:cutoff-length}. The results indicate that as the cutoff length decreases, the training performance deteriorates progressively. Given that the average length of our training data exceeds 3,000 tokens, truncation occurs when the \texttt{cutoff\_len} is set to 3,000 or less. The results show that the greater the degree of truncation to the origin data, the more significant the negative impact on the model’s final performance. Therefore, when training models for tasks involving structured data, it is crucial to ensure that the truncation length exceeds the average length of the training data to preserve training effectiveness.

\subsection{Comparison of Tool Description Length}
The length of tool descriptions plays a critical role in data synthesis, model training, and evaluation. Detailed descriptions are highly beneficial for data synthesis and generic model invocation. However, they also lead to longer context lengths, increasing GPU memory consumption and making it more challenging for the model to pay attention to key information. In our experiment, we explored the impact of workflow description length on training performance. Specifically, we categorized description lengths into three types:
\begin{compactitem}
    \item \textbf{Long Description:} The original training data containing lengthy and detailed descriptions.    
    \item \textbf{Short Descriptions:} Simplified versions of long descriptions, manually reduced to retain only essential functions and key concepts.
    \item \textbf{No Descriptions}: Tool descriptions were completely omitted.
\end{compactitem}

\begin{table}[htbp]
\centering
\caption{Examples of Tool Description with Different Length}
\label{tab:description-length}
\begin{tabular}{ll}
\toprule
Description Type &
  Example \\ \hline \vspace{-5pt}\\
Long Description &
  \begin{tabular}[c]{@{}l@{}} \{"name": "Marketing\_Employee\_Data\_Inquiry", "description": "Query the \\ name of marketing employees, year, marketing type; primary department,\\total  sales revenue, total sales revenue of the year as a percentage of the total \\revenue of the primary department, secondary department, total sales revenue\\ of the year as a percentage of the total revenue of the secondary department,\\total sales revenue of the secondary department, year; query the name of the\\ customer sold by the marketing staff, the customer code, the type of the cust-\\omer, the primary industry, the secondary industry, the Chinese brand and so\\on; query the marketing staff The name of the virtual agency, including chan-\\nel manager, organization and scoring; marketing personnel sales type setting\\ table can query the sales staff and sales type: composite, channel, customer\\and so on."\}\\ \end{tabular} \\ \vspace{-5pt}\\\hline \vspace{-5pt}\\
Short Description &
  \begin{tabular}[c]{@{}l@{}}\{"name": "Marketing\_Employee\_Data\_Inquiry", "description": "Inquire \\ about marketing employee's name, year, type of marketing, information \\about customers sold by the marketing staff, name of the virtual organization,\\ salesperson and type of sales, and much more."\} \\\end{tabular} \\\vspace{-5pt} \\\hline \vspace{-5pt}\\ 
No Description &
 \begin{tabular}[c]{@{}l@{}} \{"name": "Marketing\_Employee\_Data\_Inquiry", "description": ""\} \\\end{tabular}\\\vspace{-8pt}\\ \bottomrule
\end{tabular}
\end{table}

\begin{figure}[h]
    \centering
    \includegraphics[width=0.8\linewidth]{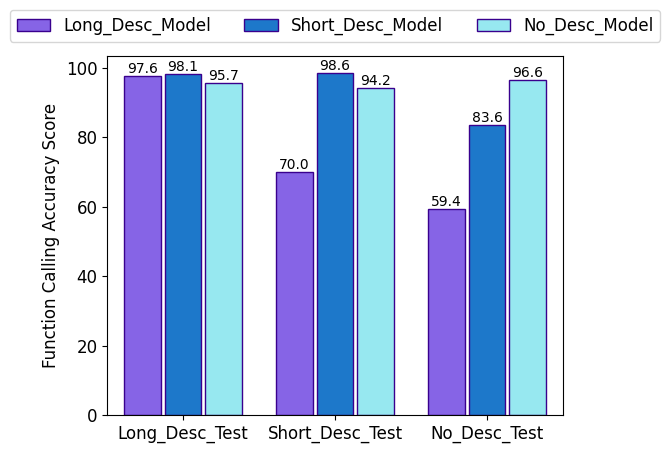}
    \caption{Comparison of Model Performance with Different Tool Description Length}
    \label{fig:description-length}
\end{figure}

Examples of different description length are shown in Table \ref{tab:description-length}. Using these variations, we constructed training and test datasets. Models were trained on their respective training sets and tested on all three test sets with different description length. Results are shown in Figure \ref{fig:description-length}. The \textbf{short description model} achieved the best performance on the short description test set, reaching an accuracy of 98.6\% on its own test set, and also demonstrated the best generalization across the long and no-description test sets. The long description model showed moderate generalization performance, while the no description model performed reasonably well across all test sets. These results suggest that shortening tool descriptions during training can effectively optimize the model’s overall performance by balancing specificity and generalization.

\newpage
\subsection{Comparison of Multi-LoRA Adapter Merging}

\begin{table}[htbp]
\centering
\caption{Impact of Merging Multiple LoRA Weights. \textit{DHR\_train\_3} is randomly split in to two training set, based on the seed data groups before augmentation. Similarly, test set is split using the same seed data groups, resulting in the \textit{DHR\_test\_A\_v1} and \textit{DHR\_test\_A\_v2} test sets. The cat (weight=1) merging method achieved the best performance.}
\label{tab:merge-lora}
\begin{tabular}{llll}
\toprule
 & \begin{tabular}[c]{@{}l@{}} \textit{DHR\_test\_A\_v1} \end{tabular} & \begin{tabular}[c]{@{}l@{}} \textit{DHR\_test\_A\_v2}\end{tabular} & Avg. \\ \hline
\rule{-3pt}{12pt}
LoRA\_a                                                                     & 94.2          & 81            & 87.6          \\
LoRA\_b                                                                     & 92.8          & 81.9          & 87.35         \\
LoRA\_linear  \textit{(weight=0.5)}      & 91.3          & 78.3          & 84.8          \\
LoRA\_linear  \textit{(weight=1.0)}        & \textbf{96.7} & 83.7          & 90.2          \\
LoRA\_dare\_linear  \textit{(density=0.8)} & 94.7          & 81            & 87.85         \\
LoRA\_cat \textit{(weight=0.5)}           & 94.7          & 82.8          & 88.75         \\
LoRA\_cat  \textit{(weight=1.0)}            & 96.6          & \textbf{84.2} & \textbf{90.4} \\
LoRA\_cat  \textit{(weight=2.0)}            & 0             & 0             & 0             \\
LoRA\_svd                                                                   & 94.7          & 82.8          & 88.75         \\
LoRA\_ties  \textit{(density=0.8)}        & 91.3          & 78.3          & 84.8          \\
LoRA\_ties\_svd  \textit{(density=0.8)}  & 94.7          & 82.8          & 88.75         \\ \bottomrule
\end{tabular}
\end{table}

To explore a training-free method, we studied the impact of merging multiple LoRA weights, aiming to enable the model to continuously incorporate new data while mitigating the effects of outdated data. Results are shown in Table \ref{tab:merge-lora}. Compared to continual learning, LoRA merging provides greater flexibility as a plug-in mechanism, allowing the model to adapt to different scenarios by selecting data from appropriate cycles. For this experiment, the \textit{DHR\_train\_3} dataset was randomly split into two parts (split by seed groups) to simulate new data generated at different time periods. These were used to train two separate LoRA models (\textbf{LoRA\_a} and \textbf{LoRA\_b}) using identical training methods. After training, we attempted various merging methods to combine the two LoRA weights and tested their effectiveness on the \textit{DHR\_test\_A} test set. The merging methods included \textbf{linear}, \textbf{cat}, \textbf{dare\_linear}, \textbf{svd}, \textbf{ties}, and \textbf{ties\_svd} \cite{peft}. The results, shown in the Table \ref{tab:merge-lora}, indicate that the cat(\textbf{weight=1}) merging method achieved the best performance. The merged LoRA outperformed both \textbf{LoRA\_a} and \textbf{LoRA\_b}, demonstrating superior generalization capabilities. This highlights the potential of LoRA merging not only to improve model adaptability but also to provide a flexible and efficient approach for managing data life-cycle effects in enterprise scenarios.

\section{Conclusion}
In this paper, we introduced a specialized training pipeline for function-calling capabilities in professional scenarios. This pipeline encompasses the generation and augmentation of scenario-specific function-calling data, LoRA fine-tuning of models, and comprehensive evaluation and analysis of function-calling performance. The experimental results demonstrate that our pipeline can improving the accuracy of generic function-calling models in specialized scenarios. Notably, a 7B-parameter model trained using this pipeline achieved accuracy levels superior to state-of-the-art LLMs. This is particularly significant for enterprises, as the pipeline enables the automated training of intelligent agents' core engines tailored to various business scenarios. 

Beyond the digital HR scenario explored in this study, the pipeline can be applied to other contexts, such as integrating with edge models in edge devices, supporting diagnostic systems in healthcare, or powering super employees within organizations. Furthermore, with the inclusion of a data feedback module, the system could utilize data generated from users' interactions with agent applications. By leveraging AI and the designed pipeline, the returning data can be automatically annotated and used for continual model training, enabling iterative updates.

\section{Limitation \& Future Work}
Currently, the proposed pipeline has certain limitations. First, it focuses solely on single-turn data synthesis and does not account for more complex scenarios, such as multi-step and multi-turn interactions, which are highly practical in real-world applications. Additionally, the current pipeline's data filtering process for generated data is relatively simplistic. Future efforts will aim to enhance the filtering of augmented data to further improve data quality. Despite these limitations, we believe this pipeline represents a step forward in applying LLMs to real-world enterprise scenarios. 

\printbibliography

%%%%%%%%%%%%%%%%%%%%%%%%%%%%%%%%%%%%%%%%%%%%%%%%%%%%%%%%%%%%

\newpage
\appendix
\renewcommand{\thesection}{Appendix \Alph{section}}

\section{Data Synthesis Prompt Templates} \label{sec:apx_a}
\textit {Prompt Templates are translated to English.}
\subsection{Question Generation Prompt Template}
\begin{lstlisting}
    prompt=f"""You will act as a user asking questions, using the provided function parameter list {key}:{params_data} to generate questions. 
    Here, {key} represents the function name, and the function description is {function_description}. Follow the requirements below:     
        - Generate {number} questions for each function.   
        - Include names, ID strings, and ID parameters in each question, ensuring that the ID names are retained.   
        - When the question involves employee IDs, you may randomly generate human names.   
        - The ID is known information and must be shown in the question.   
        - Do not include the function name in the question.   
        - Only generate the questions, without answers or any additional content.    
    Output format example:   
    1. Question content   
    2. Question content   
    ...   
    5. Question content"""
\end{lstlisting}

\subsection{Question Generation with real name}
\begin{lstlisting}
    prompt=f"""You are a data annotator. Your task is to generate a set of diverse questions for the given function.   
    The constructed questions will be directly used as parameters for function calls, demonstrating how these functions can be applied in real-world scenarios.    
    
    Please adhere to the following requirements:   
    - Ensure the questions are as diverse as possible and not limited to the "example" format.   
    - Project names, course names, etc., can be reasonably fabricated.    
    
    Ensure the following:   
    - Provide a variety of questioning styles and tones, consistent with those used by various types of consulting professionals.   
    - The first half of the questions should consist of **{half} single queries** (e.g., "xxxxxx?"), and the second half should consist of **{half} multi-part queries** (e.g., "xxxxxx? xxxxxx? xxxxxx?").   
    - Generate questions based on the function tool's description and usage but avoid explicitly mentioning the function name.   
    - Include characters (IDs or names) in the questions, ensuring that all IDs or names are selected from the following list: **{ids}**.   
    - Generate each question on a single line. Do not include any bullet points, numbering, or blank lines. The response should contain only the generated questions, without any additional content.    
    
    Based on the above instructions and examples, generate **{number} different questions** for the function **"{func_name}"**.    
    The detailed function description is as follows:   **{func_desc}**    
    
    Now, generate a total of **{number} different questions** following the format above. Ensure the number of questions is accurate. """
\end{lstlisting}

\subsection{Question Generation without name}
\begin{lstlisting}
    prompt=f"""You are a data annotator. Your task is to generate a set of diverse questions for the given function.   
    The constructed questions will be directly used as parameters for function calls, demonstrating how these functions can be applied in real-world scenarios.    
    
    Please adhere to the following requirements:   
    - Ensure the questions are as diverse as possible and not limited to the "example" format.   
    - Project names, course names, etc., can be reasonably fabricated.    
    
    Ensure the following:   
    - Provide a variety of questioning styles and tones, consistent with those used by various types of consulting professionals.   
    - The first half of the questions should consist of **{half} single queries** (e.g., "xxxxxx?"), and the second half should consist of **{half} multi-part queries** (e.g., "xxxxxx? xxxxxx? xxxxxx?").   
    - Generate questions based on the function tool's description and usage but avoid explicitly mentioning the function name.   
    - Avoid including specific individuals in the questions; instead, focus on companies or groups of people.   
    - Generate each question on a single line. Do not include any bullet points, numbering, or blank lines. The response should contain only the generated questions, without any additional content.    
    
    Based on the above instructions and examples, generate **{number} different questions** for the function **"{func_name}"**.    
    
    The detailed function description is as follows:   **{func_desc}**    
    
    Now, generate a total of **{number} different questions** following the format above. Ensure the number of questions is accurate. 
    """
\end{lstlisting}

\subsection{Short Question Generation}
\begin{lstlisting}
    prompt=f"""You are a data annotator. Your task is to perform data augmentation on the given question based on the function description, expanding the phrasing of the question.    
    
    Question to be expanded: "{question}"    
    
    Please follow these guidelines:   
    - Use various modification methods, including but not limited to changing the order of elements, question style, tone, and omissions.   
    - You may expand or simplify the question.   
    - Subject, verb, and object can be rearranged, such as placing names at the end of the sentence.   
    - If the question includes characters, you may replace the names in the enhanced versions with names or IDs from the following list: {ids}.    
    
    Ensure that the original meaning of the question remains unchanged.   
    Ensure that the enhanced versions are still short questions.  
    Ensure that only the final {number} answers are output, with each question on a separate line. Do not include additional content or numbering.    
     
    Based on the examples and instructions above, expand the question into a total of {number} different versions. Ensure the correct number of outputs. 
    """
\end{lstlisting}

\section{Data Augmentation Prompt Templates}  \label{sec:apx_b}
\subsection{Question Replacement}
\begin{lstlisting}
    prompt = f""" You are now a data annotation expert capable of rewriting questions according to specific requirements. Note: Directly respond with the rewritten questions. Generate only the rewritten answers without any extra content or numbered bullet points.    
    
    Please follow these instructions:   
    - Carefully review the contents in **{info_dict}**, and ensure that the rewritten questions align with the descriptions provided.   
    - Replace key information and content, such as names, dates, various labels, locations, etc.   
    - Replace any names appearing in the question, e.g., Wang XX, with other names from the following list: **{names}**. Ensure all names are replaced.   
    - Replace any departments appearing in the question, e.g., Marketing Department, with other departments from the following list: **{departments}**.   
    - Replace any cities appearing in the question, e.g., Beijing, with other cities from the following list: **{cities}**.   
    - Replace years mentioned in the question with one of the following: **2021, 2022, 2023**.   
    - Replace any other numbers mentioned, e.g., Issue 344, with random numbers.   
    - Replace any zodiac signs, e.g., Sagittarius, with others such as Capricorn, Leo, Taurus, Virgo, Aquarius, etc.   
    - Replace any project amounts mentioned with random values, e.g., 100K, 500K, 2M, etc.   
    - Replace any job positions, e.g., Product Manager, with others such as Data Analyst, Marketing Specialist, Finance Manager, HR Specialist, General Manager, etc.   
    - Replace any skills mentioned, e.g., Big Data Development, with others such as Software Development, Foreign Language Proficiency, Project Management, Organizational Coordination, Planning Skills, etc.   
    - Replace any technologies mentioned, e.g., Deep Learning, with others such as Data Governance, Large Model Fine-Tuning, Prompt Engineering, Machine Learning, Computer Vision, etc.   
    - Ensure the rewritten questions remain simple and focus on core information. Avoid using prepositions, conjunctions, modal particles, or honorifics. Do not include quotation marks.    
    
    Rewrite the question **"{question}"** in **{number} different ways**.   
    """
\end{lstlisting}

\subsection{Question Rewriting}
\begin{lstlisting}
    prompt = f"""You are now a linguist capable of rewriting user queries. Your task is to change the grammatical structure and questioning style of the sentence while preserving its original meaning and key information. Key information is referenced from {info_dict}.    
    Please rewrite the given question into {number} different forms, altering the syntax and questioning style without adding new information or omitting any original content.    
    - Each rewritten question should be on a separate line.   
    - Do not include any additional content, quotation marks, or numbering.   
    - The rewritten questions will be directly used as parameters for function calls, showcasing how these functions are applied in real-world scenarios.    
    
    Rewrite the following question:   
    {question}   
    """
\end{lstlisting}

\subsection{Question Simplification}
\begin{lstlisting}
    prompt = f"""You are a data annotator. Your task is to simplify the given question based on the tool function description. The question you need to simplify is: "{question}".    
    
    Please adhere to the following requirements:   
    - Simplify the question into a very concise phrase.   
    - Do not change the original meaning of the question.    
    
    Example:   
    Original question: "What is Zhang San's current salary?"   
    Simplified: "Zhang San salary"    
    
    Provide one question per line. Do not include any additional content, quotation marks, or bullet points.    
    Based on these examples and the instructions above, generate a total of {number} different simplified questions. Ensure the number of questions is correct.
    """ 
\end{lstlisting}

\subsection{Question Error Introduction}
\begin{lstlisting}
    prompt = f"""You are now a data annotator tasked with simulating user input errors by modifying the original query. Make the following types of changes:   
    - rewrite the query "{question}" to produce **{number} rewritten versions**.    
    - Replace one character in the user query with a phonetically similar typo.  
    - Replace some words with incomplete or scrambled versions of the words.   
    - Repeat some characters within the query.    
    - Respond directly with the rewritten queries and do not include any extra content.   
    - Provide each rewritten query on a separate line.   
    - Do not include quotation marks, numbering, or any additional information in your response.   
    - Ensure the number of rewritten queries matches the requested count.    
    
    Example: 
    Original query: "How many technical staff members are in the company?" 
    Rewritten: "How many technical stuff members are there in most companie?"   
    
    Follow these examples and instructions to rewrite the query. Ensure the quantity of responses is accurate. 
    """
\end{lstlisting}

%%%%%%%%%%%%%%%%%%%%%%%%%%%%%%%%%%%%%%%%%%%%%%%%%%%%%%%%%%%%

\end{document}